\def\year{2019}\relax
\documentclass[letterpaper]{article} 
\usepackage{aaai19}  
\usepackage{times}  
\usepackage{helvet}  
\usepackage{courier}  
\usepackage{url}  
\usepackage{graphicx}  

\begin{document}
%

\title{Multimodal Deep Neural Networks using Both Engineered and Learned Representations for Biodegradability Prediction}

\author{Garrett B. Goh,\textsuperscript{1,*}
Khushmeen Sakloth,\textsuperscript{2}
Charles Siegel,\textsuperscript{1}
Abhinav Vishnu,\textsuperscript{1}
Jim Pfaendtner, \textsuperscript{2}\\
\textsuperscript{1}Pacific Northwest National Lab (PNNL), \textsuperscript{2}University of Washington (UW)\\
\underline{garrett.goh@pnnl.gov}, ksakloth@gmail.com, charles.siegel@pnnl.gov, abhinav.vishnu@pnnl.gov, jpfaendt@uw.edu
}

\maketitle
\begin{abstract}
Deep learning algorithms excel at extracting patterns from raw data, and with large datasets, they have been very successful in computer vision and natural language applications. However, in other domains, large datasets on which to learn representations from may not exist. In this work, we develop a novel multimodal CNN-MLP neural network architecture that utilizes both domain-specific feature engineering as well as learned representations from raw data. We illustrate the effectiveness of such network designs in the chemical sciences, for predicting biodegradability. DeepBioD, a multimodal CNN-MLP network is more accurate than either standalone network designs, and achieves an error classification rate of 0.125 that is 27\% lower than the current state-of-the-art. Thus, our work indicates that combining traditional feature engineering with representation learning can be effective, particularly in situations where labeled data is limited. 
\end{abstract}

\section{Introduction}
\label{sec:intro}

Despite decades of research, designing chemicals with specific properties or characteristics is still heavily driven by serendipity and chemical intuition. Current machine learning (ML) and deep learning (DL) models in chemistry typically rely on engineered features (molecular descriptors) developed from expert knowledge, and have made progress in property prediction.

\subsection{Big Data but Small Labels}
Like many other fields, the growth of big data in chemistry is underway~\cite{goh2017r}. However, the amount of labeled data is significantly smaller than that available in conventional DL research. To illustrate this disparity, a database of 100,000 measured (labeled) samples is considered a significant accomplishment in chemistry, which would otherwise be considered a small dataset in computer vision research, where datasets such as ImageNet that includes over a million images are typically the starting point. Owing to the complexities of data collection in the chemical sciences which require expensive and slow wet-lab experimentation, the scarcity of labels is an inherent problem to this field. 

While representation learning thrives in a big data environment, recent DL models that learn directly from raw chemical data have accomplished reasonable success using molecular graphs,~\cite{duvenaud2015,kearnes2016} molecular images,~\cite{goh2017c1,goh2017c2,wallach2015}, or molecular text-representations~\cite{bjerrum2017,goh2017s} to predict chemical properties. However, the watershed breakthrough akin to AlexNet for computer vision has yet to be observed in this field, although recent developments in weak supervised training methods such as ChemNet has made progress towards this goal.~\cite{goh2017c3} \textit{Therefore, while big data in chemistry exists, it comes with a caveat of small labels, which reduces the effectiveness of deploying deep learning in this industry.}

\subsection{Contributions}
Our work addresses some of the challenges associated with the big data but small label challenge in chemistry. \textit{Specifically, we develop a multimodal CNN-MLP network architecture that incorporates both engineered features and learned representations, which is the first reported example of how multimodal learning can be used effectively in chemistry}. Our contributions are as follows. 
\begin{itemize}
	\item We develop the first multimodal CNN-MLP neural network architecture for chemical property prediction that utilizes both engineered and learned representations.	
	\item We investigate the effect of network architecture, hyperparameters, and feature selection on model accuracy.
	\item We demonstrate the effectiveness of this network design in the development of DeepBioD, which considerably outperforms existing state-of-the-art models by 31\% for predicting chemical biodegradability.
\end{itemize}

The organization for the rest of the paper is as follows. In section 2, we outline the motivations and design principles behind the multimodal network. In section 3, we examine the biodegradability dataset, its applicability to chemical-affiliated industries, and the training protocol. Lastly, in section 4, we explore different multimodal network designs, and other factors that affect model accuracy and generalization. We conclude with the development of the DeepBioD model for predicting biodegradability, and evaluate its performance against the current state-of-the-art.

\subsection{Related Work}
Multimodal learning is an established technique in DL research. Our work takes inspiration from earlier research that demonstrated using different data modalities can improve model accuracy~\cite{ngiam2011}. However, to the best of our knowledge existing multimodal learning models operate primarily on different streams of synchronous raw data, for example a video stream and its corresponding audio stream, or an image and its respective text caption. In contrast, there has been limited research in using multimodal learning to combine traditional feature engineering with representation learning, and there currently exist no examples of multimodal learning in chemistry.

\section{Multimodal Network Design}
\label{sec:design}

In this section, we document the data preparation steps for processing chemical image data. Then, we examine the design principles behind the multimodal neural network used in this work.

\subsection{Data Representation}

We followed the same preparation of chemical image data as reported by Goh et. al.~\cite{goh2017c1}. Briefly, the 2D molecular structure and its coordinates were used to map onto a discretized image of 80 x 80 pixels that corresponds to 0.5 {\AA} resolution per pixel. Then, each atom and bond pixel is assigned a "color" using the "EngD" color-coding scheme as reported by Goh et. al.~\cite{goh2017c2}. The resulting image is then used to train the Chemception CNN model.

In addition to the chemical image data, the other component of the multimodal model, the MLP network, uses engineered features (molecular descriptors) as input. Two sets of molecular descriptors input data were obtained. The first set, referred to as Ballabio-40, was obtained directly from Mansouri et. al.~\cite{mansouri2013}, which is a set of 41 selected descriptors that was computed from DRAGON. We also prepared a second set, PaDEL-1400, which is a more comprehensive set of \textasciitilde1400 descriptors computed using PaDEL.

\subsection{Designing the Multimodal Neural Network}

\begin{figure}[!htbp]
\centering
\includegraphics[scale=0.10]{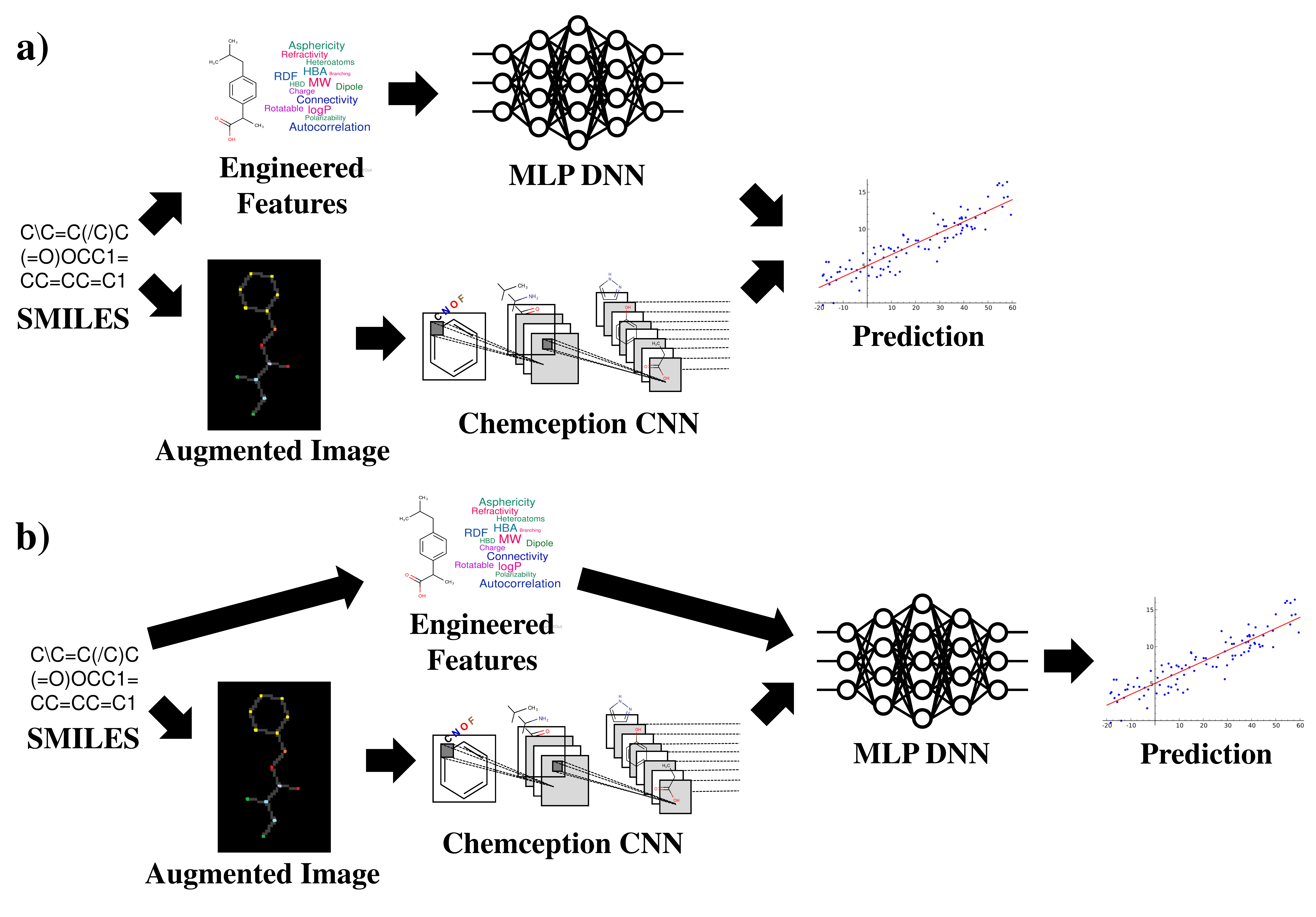}
\caption{\small Schematic diagram of the multimodal neural network that uses both molecular descriptors and raw image data that operates in (a) parallel and (b) sequential mode.}
\label{fig:schematic}
\end{figure}

CNNs are effective neural network designs for learning from image data. Its effectiveness has been demonstrated in examples like ResNet~\cite{he2015} that achieves human-level accuracy in image recognition tasks. In the absence of copious amount of data, the representation learning ability of deep neural networks may not learn optimal features. In the context of limited labeled chemical data, any sub-optimality of the network's learned representations could potentially be mitigated with the introduction of engineered chemical features. Thus, the goal of this work is to combine two modalities, the first is engineered features, and the other is using raw image data. However, unlike conventional multimodal DL models, there is no synchronization between the data streams in this work. Therefore, an appropriate multimodal network design that works well for chemistry research problems must be first developed.

As illustrated in Figure~\ref{fig:schematic}, we explored two multimodal architectures that operate as either a parallel or sequential model. In the parallel model, two neural networks are trained simultaneously. The first component is a standard feedforward MLP neural network that uses molecular descriptors as the input data. The other is the Chemception CNN model that uses chemical image input data. The penultimate layer output from the CNN will correspond to learned features of the entire chemical. Using expert knowledge, we know this output is similar to a molecular descriptor (i.e. engineered features of the entire chemical). We therefore, concatenated both network outputs before passing it to the final classification layer. This approach thus recombines the learned features from the CNN with the reprocessed features from the MLP (we use the term "reprocessed" as the output from the MLP will be a non-linear combination of the original input of engineered features). The underlying hypothesis of this design is that during training, the CNN component could potentially learn representations to supplement missing features from the MLP component.

An alternative approach would be to train a multimodal model in two stages. In this setup, known as the sequential model, the Chemception CNN model is first trained directly on the chemical task. Once this is completed, the CNN weights are freezed, and the penultimate layer output is concatenated with the molecular descriptors, which are then collectively used as input data for training a second MLP network. This approach is therefore equivalent to running a list of engineered features (molecular descriptors) and learned representations (from the CNN) through a MLP network. The underlying hypothesis behind this design is that the CNN is allowed to first learn its own representations, and any inherent shortcomings in either its learned features or expert engineered features may be mitigated through additional representation learning in the MLP network.

As for the network architecture of the Chemception CNN model, we used the T3\_F16 model reported by Goh et. al.~\cite{goh2017c1}. For the MLP architecture, we performed a grid search totaling 20 different versions of each multimodal network design for [2,3,4,5] fully-connected layers with ReLU activation functions~\cite{glorot2011} and [16,32,64,128,256] neurons per layer. A dropout of 0.5 was also added after each fully-connected layer. The results presented in this paper for each respective multimodal model is the best model found using grid search, as determined by the validation error rate.

\section{Methods}
\label{sec:method}

In this section, we provide an introduction to the biodegradation dataset. Then, we document the training methods, as well as the evaluation metrics used in this work.

\subsection{Industrial Applications (Biodegradability)}

One example of a chemistry research problem with limited data is in biodegradability prediction. Biodegradability is the tendency of chemicals to break down naturally. Non-biodegradable chemicals that do not decay, can lead to accumulation and this can be harmful in the long-term for the environment. Furthermore, biodegradation is a long-time scale process, and as such, using ML/DL models to predict biodegradability is a rapid and cost-effective solution. Over the years, ML models have been developed and the current state-of-the-art is based on conventional ML algorithms trained on engineered features (molecular descriptors).~\cite{mansouri2013}

\subsection{Dataset Description}

In this work, we used the same dataset that was used to develop the current state-of-the-art model for predicting biodegradability~\cite{mansouri2013}. Each chemical is classified as either biodegradable (RB) or non-biodegradable (NRB). The training/validation dataset was curated from MITI. The test set was constructed from two data sources, Cheng et.al. and the Canadian DSL database. Additional data cleaning steps, such as handling data replicates, unifying test duration, etc. were reported previously, and we used the final cleaned dataset for training DeepBioD and benchmarking against existing models~\cite{mansouri2013}.

\begin{table}[!htbp] 
		\centering
		\begin{tabular}{|c|c|c|c|}
				\hline
				Data & RB & NRB & Total \\
				\hline 
				Training/Validation & 356 & 699 & 1055\\
				Test & 191 & 479 & 670\\
				\hline
		\end{tabular}\\  
		\caption{Biodegradability dataset used in this study.}
		\label{table:dataset}  
\end{table}

\subsection{Data Splitting}

We used a random 5-fold cross validation for training and evaluated the performance and early stopping criterion of the model using the validation set. The splitting between training/validation and test dataset was identical to Mansouri et. al.~\cite{mansouri2013}, and it was used throughout this paper unless specified otherwise.

We also noted that the training/validation and test datasets were obtained from different sources. This means that the chemicals between both datasets may not overlap well in chemical space, and/or systematic biases from different experimental measurements or lab protocols might arise. Thus, we also preprocessed a re-mixed dataset to mitigate the above-mentioned effects. In this re-mixed dataset, both training/validation and test datasets were combined, and a random 40\% was re-partitioned out to construct a new test set. The re-mixed training/validation and test dataset is therefore like the original datasets in terms of size, but each dataset would have samples from all data sources.

\subsection{Training the Neural Network}

DeepBioD was trained using a Tensorflow backend~\cite{abadi2016} with GPU acceleration. The network was created and executed using the Keras 1.2 functional API interface~\cite{chollet2015}. We use the RMSprop algorithm to train for 500 epochs using the standard settings recommended: learning rate = 10\textsuperscript{-3}, $\rho$ = 0.9, $\epsilon$ = 10\textsuperscript{-8}. We used a batch size of 32, and also used early stopping to reduce overfitting. This was done by monitoring the loss of the validation set, and if there was no improvement after 50 epochs, the last best model was saved as the final model. During the training of the Chemception CNN component, we also performed additional real-time data augmentation to the image using the ImageDataGenerator function in the Keras API, where each image was randomly rotated between 0 to 180 degrees.

\subsection{Loss Function and Performance Metrics}

We used the binary cross-entropy loss function for training. The performance metric reported in our work is the classification error rate (Er), which was defined in prior publications and it is a function of sensitivity (Sn) and specificity (Sp):

\[Er  =  1 - (Sp - Sn)/2\]
\[Sp = \frac{TN}{TN + FP}\quad Sn = \frac{TP}{TP + FN}\]

\section{Experiments}
\label{sec:exp} 

In this section, we perform several experiments to determine the best multimodal network design, which is used to develop DeepBioD, an ensemble model for predicting biodegradability. Then, we compared DeepBioD to existing state-of-the-art methods for biodegradability prediction.

\subsection{Searching for an Optimal Network Design}
We investigated factors that affected the performance of the multimodal neural network. In the absence of more data, network architecture has been a key driver in increasing model accuracy~\cite{he2015}. We also investigated the effect of feature selection, and examine an alternative data splitting approach to account for systematic biases.

\subsubsection{Evaluating Baseline Models}
As our work combines a typical MLP network with the Chemception CNN model, we first evaluated the performance of baseline (single-modal) models. Two different training strategies were used for Chemception. The first approach was supervised learning on the biodegradability dataset. The second approach is based on ChemNet~\cite{goh2017c3}, which is a weak supervised transfer learning approach that uses a model pre-trained on numerous chemical rules and fine-tuning its weights for predicting biodegradability. The use of ChemNet in this context is therefore analogous to using existing image classification models (ResNet, GoogleNet, etc.) for fine-tuning on related tasks.

\begin{figure}[!htbp]
\centering
\includegraphics[scale=0.20]{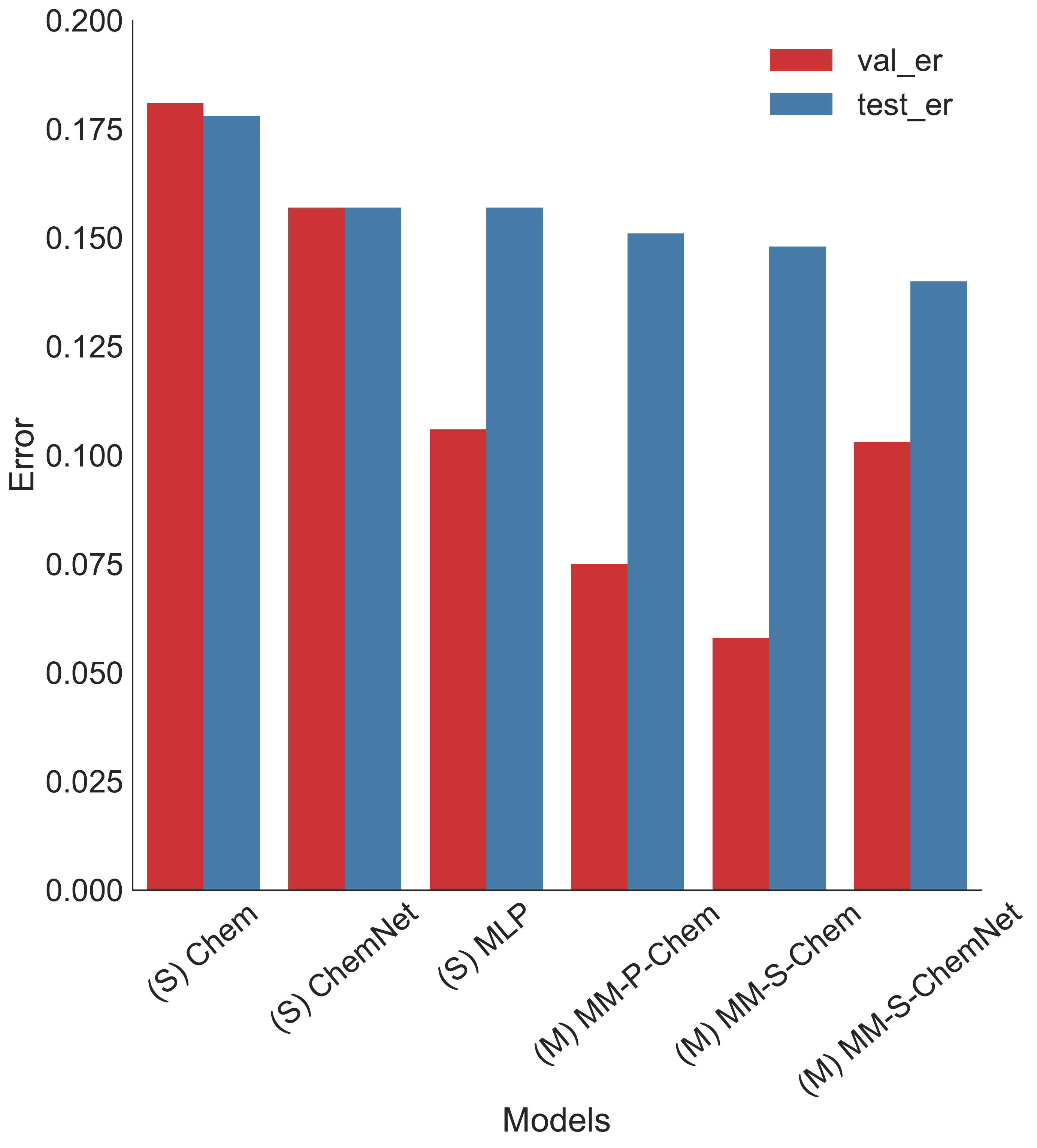}
\caption{\small Error classification rate of standalone (S) Chemception or MLP models compared to multimodal (M) CNN-MLP models, trained on the Ballabio-40 descriptor set.}
\label{fig:standalone_error}
\end{figure}

As shown in Figure~\ref{fig:standalone_error}, Chemception trained directly on the biodegradability dataset achieved a classification error rate of 0.178 on the test set. Relative to traditional ML algorithms trained on the same dataset, which achieved an error rate of 0.170 to 0.180~\cite{mansouri2013}, the resulting Chemception model is comparable, but it is not performing better. We attribute this observation to a consequence of having limited labeled data. In comparison, ChemNet achieved a noticeably lower error rate of 0.157. This model exploits the advantage of weak supervision that uses a much larger database of 500,000 compounds, but without using additional labels. In addition, the baseline MLP model that was trained on the Ballabio-40 descriptor set also achieved a similar error rate of 0.156. \textit{Our results indicate that neither a standalone MLP nor CNN model is better than the other.}

\subsubsection{Parallel vs Sequential Multimodal Network}

Having established baseline standards, we now explore if the inclusion of engineered features in a multimodal configuration will improve model performance. We trained both sequential and parallel multimodal neural networks. For the parallel model (MM-P-Chem), the Chemception component was trained directly on the dataset. For the sequential model, we used the fixed weights of either the Chemception (MM-S-Chem) or ChemNet (MM-S-ChemNet) model trained in the preceding section, to generate additional input that was passed to the MLP network. As shown in Figure~\ref{fig:standalone_error}, all multimodal models perform better than the standalone MLP or Chemception models.

We observed the error rate of both MM-P-Chem (0.151) and MM-S-Chem (0.148) models trained directly on the Ballabio-40 dataset are similar. However, the large difference between validation and test error rates suggest that the multimodal models are not generalizing as well as the standalone models. There are 2 factors that account for this observation. First, by virtue of limited labels, there may be insufficient diversity being presented in the training/validation set, which may not overlap well with the test set (i.e. the training/validation and test datasets are not sufficiently similar). The other factor to consider is that the test dataset was obtained from two separate database/measurements, which may introduce additional systematic biases.

To improve the generalization of the multimodal models, we evaluated the effect of using the ChemNet model. This model was originally trained on a much larger and more diverse dataset, before fine-tuning on the biodegradability dataset. We tested both parallel and sequential models but observed that only the sequential multimodal model could be trained effectively. We suspect this is because the parallel model uses a pre-trained ChemNet for the CNN component, but the MLP component would be initialized with random weights. It is plausible that more sophisticated training methods can be developed, but it is beyond the scope of this work.

The results presented in Figure 2, shows that the MM-S-ChemNet model achieved a validation and test error rate of 0.103 and 0.140 respectively. Compared to the Chemception-based multimodal models, the difference in error rate is smaller, but more importantly, the test error rate is also the lowest amongst all models explored. \textit{Therefore, these results strongly indicate that multimodal models that utilize both raw data and engineered features can improve model accuracy relative to standalone models, and limitations in generalizing to new data can be mitigated by pre-training on larger datasets.}

\subsubsection{Gaining Insight on Feature Selection}

The performance of the MLP and multimodal models reported thus far have trained on the selected set of 41 molecular descriptors that were used to develop the current state-of-the-art model~\cite{mansouri2013}. In that work, the authors constructed the Ballabio-40 descriptor set from a larger pool of 800 descriptors, using a sophisticated feature selection protocol based on clustering and genetic algorithms to identify the optimal subset of molecular descriptors for training conventional ML models.

\begin{figure}[!htbp]
\centering
\includegraphics[scale=0.45]{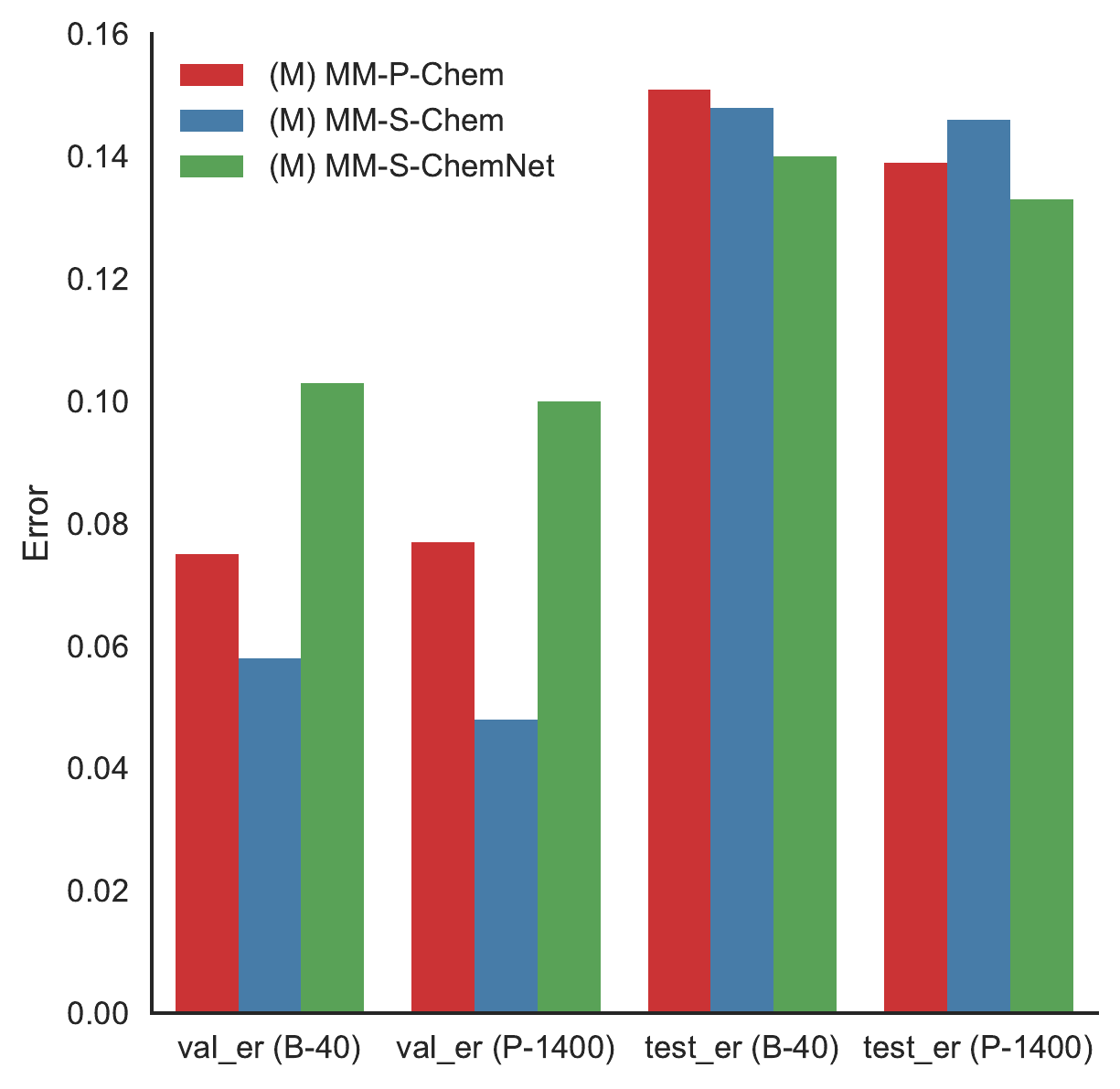}
\caption{\small Error classification rate of various multimodal CNN-MLP models when trained on the PaDEL-1400 descriptor set shows a systematic reduction relative to comparable models trained on the Ballabio-40 descriptor set.}
\label{fig:multimodal_error}
\end{figure}

Unlike traditional ML algorithms, deep neural networks with modern algorithms and training methods have been shown to be robust even in the presence of many input features,~\cite{lecun2015} and consequently feature selection may not be necessary. In addition, excessive feature selection has the effect of reducing the quantity of input data to the MLP network, and this means that the network has less data to work with in learning representations, which may degrade its performance. Hence, we explored the effect of using a larger set of ~1400 features computed using PaDEL. We performed analogous experiments, and the results are summarized in Figure~\ref{fig:multimodal_error}. We observed a systematic reduction in the test error rate across all 3 multimodal models evaluated when using the PaDEL-1400 descriptor set. In addition, the best model, MM-S-ChemNet, has its error reduced from 0.140 to 0.133, which is the lowest error attained thus far. \textit{These results indicate that excessive feature selection may not be necessary when using deep neural networks, and the inclusion of more engineered features can improve model performance.}

\subsubsection{On Dataset Dissimilarity and Systematic Biases}

With limited labeled data, there is a risk that training/validation and test datasets may not be sufficiently similar, and if datasets are constructed from different sources, such as in this work, this may introduce systematic biases due to differences in experimental techniques and protocol. To further disentangle our model's accuracy from these effects, we explore a different data splitting protocol. As detailed in the methods section, we reconstructed a new training/validation and test dataset that had samples from all 3 sources used to construct the original biodegradability dataset.


This re-mixed dataset was trained with the PaDEL-1400 descriptor set. We observed that the validation error showed no systematic improvement across various multimodal models tested. However, there was a consistent reduction in the test error rate, to the point that the best model, MM-S-ChemNet achieves an error rate of 0.114, compared to 0.133 using the original data splitting. This suggests that the difference in the two error rates can be attributed to any dissimilarity between the original datasets, as well as any systematic biases introduced by different databases/experiments. \textit{Our results suggest that when developing models with limited training data, models should be periodically retrained with new data (when available) to improve generalizability, and to account for unknown biases (if any).}

\subsubsection{DeepBioD: Putting it All Together}

Having explored various multimodal model network designs, and the factors that affect model performance, we have identified that the MM-S-ChemNet model trained on a larger PaDEL-1400 descriptor set, where the MLP component has 2 fully-connected layers of 128 neurons each, provides the lowest error and best generalization. To develop DeepBioD, we trained a total of 5 additional MM-S-ChemNet models that use a different seed number to govern the splitting between training and validation datasets. These 5 individual models were then used to develop an ensemble model, in which the predicted output across all 5 models was averaged, and the mean output was used to predict the molecule's class. \textit{Using this ensemble approach, the final DeepBioD model achieves an error classification rate of 0.125 on the test set.}

\subsection{Comparing DeepBioD to State-of-the-Art Models}

To the best of our knowledge, no DL model has been developed for biodegradability prediction. The current state-of-the-art is a consensus model of 3 traditional ML algorithms: kNN, PLSDA, SVM~\cite{mansouri2013}. In addition, two type of consensus models were reported. The consensus \#1 model uses the average output of the 3 individual ML algorithms to provide a final prediction, and this is similar in approach to our ensemble model.

\begin{figure}[!htbp]
\centering
\includegraphics[scale=0.40]{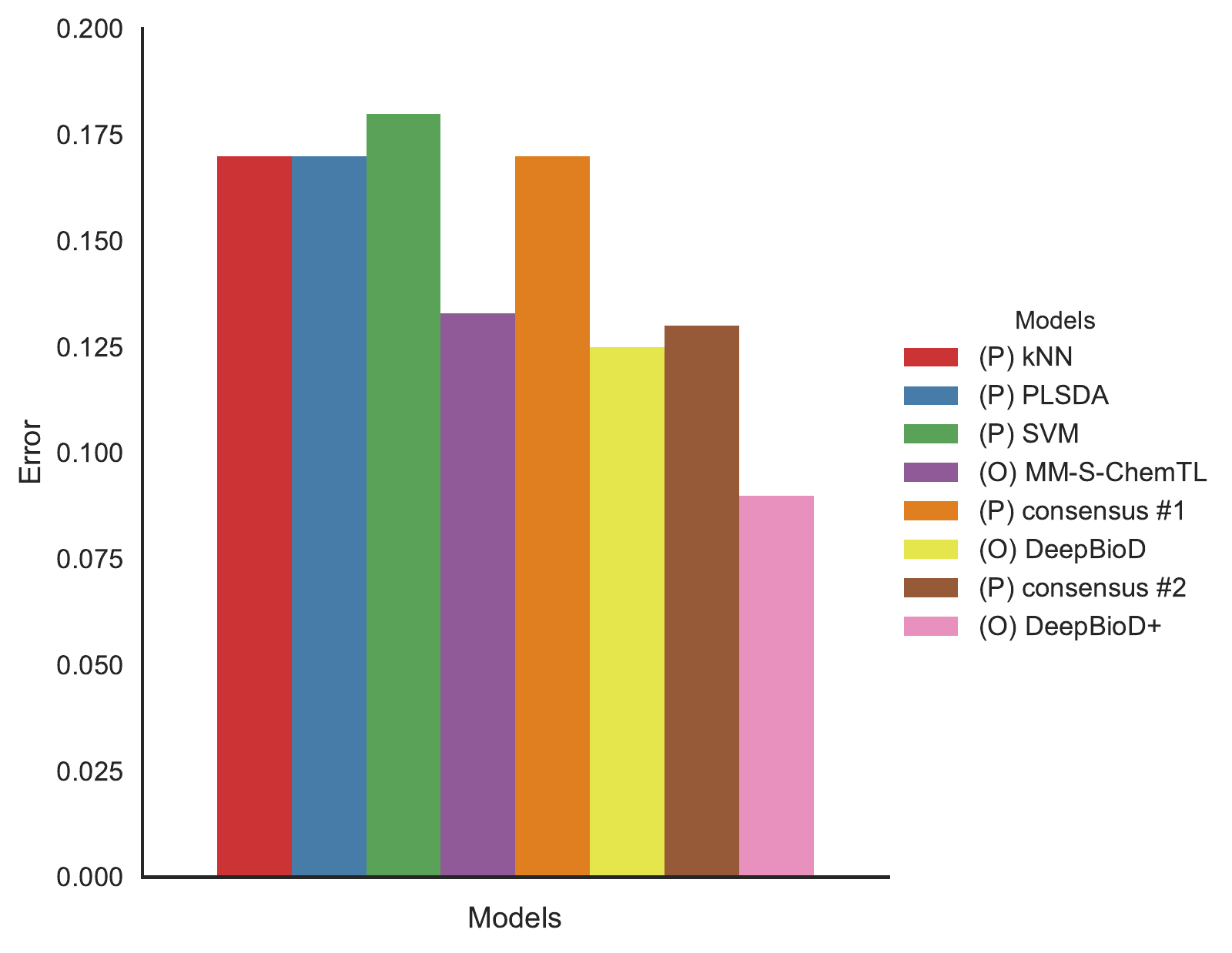}
\caption{\small Our model (O) DeepBioD has a lower error classification rate compared to prior (P) state-of-the-art models for biodegradability predictions in every category.}
\label{fig:deepbiod}
\end{figure}

As illustrated in Figure~\ref{fig:deepbiod}, a single multimodal neural network model, MM-S-ChemNet that has an error rate of 0.133 already outperforms all 3 ML models, as well as the consensus \#1 model that has an error rate of 0.170. However, a more appropriate comparison against consensus \#1 would be using an ensemble model like DeepBioD that achieves an error rate of 0.125. \textit{Therefore, DeepBioD provides a 27\% reduction in error rate relative to current state-of-the-art.}

\subsubsection{Reliability and Generalizability Tradeoff}

When dealing with limited data, it is important to ascertain the reliability of a model's prediction. Prior work has demonstrated that using the consensus \#2 model, which only returns a predicted class when all 3 underlying ML models are in agreement can reduce the error rate to 0.130~\cite{mansouri2013}. However, the drawback is that this model is only able to provide predictions for 87\% of the test set. It has to be emphasized that DeepBioD in its current form provides a classification for all the compounds in the test set, and with a lower error rate. However, if one were to factor in the completeness of prediction coverage, a more appropriate comparison to consensus \#2 would be a modified model that is adjusted to return no classes when its prediction is not reliable. In this modified DeepBioD+ model, we introduced an empirical threshold criterion that filters out predictions if the class probability is less than 0.8. The resulting model achieves an error rate of 0.090 with 89\% coverage, which is a non-trivial 31\% reduction relative to the current state-of-the-art consensus \#2 model. \textit{These results indicate that filtering out less reliable predictions can boost model performance, which can be particularly useful in prospective studies on unseen data.}

\subsection{Multimodal Learning in Other Domains}

While the multimodal model that we have developed is for biodegradability prediction, this approach can be extended to other properties of interest to chemical-affiliated industries, on the condition that some labeled data exists. We also anticipate that the CNN component can be replaced with different network architectures, such as those that use graphs ~\cite{duvenaud2015,kearnes2016} or text ~\cite{bjerrum2017,goh2017s} as the input data.

Lastly, there are design principles that can be adapted to other domains. First, (i) prior feature engineering research by domain scientists is necessary, which implies that other scientific, engineering and financial modeling applications may benefit from this approach. Second, (ii) identifying appropriate locations to combine data streams from engineered and learned representations is critical. In our work, we used expert-knowledge to determine that the penultimate layer output of the CNN component would correspond to engineered features. For other domain applications, we anticipate this solution will be field specific requiring both expert-knowledge and deep learning ingenuity to identify similar critical points in the multimodal network design.

\section{Conclusions}
\label{sec:conclusions}

In conclusion, we developed a novel multimodal CNN-MLP neural network architecture for predicting biodegradability. This model uses both raw data (images) and engineered features (molecular descriptors), while leveraging weak supervised learning and transfer learning methods. The final DeepBioD model achieves an error rate of 0.125, significantly outperforming the current state-of-the-art of 0.170. We have also shown that training on larger datasets using weak supervision (without requiring additional labels) improves the model's ability to generalize to new data. Our work therefore demonstrates that a multimodal network that combines the benefit of representation learning from raw data with expert-driven feature engineering is a viable approach in domain applications that have limited labeled data.

\bibliographystyle{aaai}
\bibliography{references}

\end{document}